\documentclass{article}
\usepackage{colortbl}
\usepackage{microtype}
\usepackage{graphicx}
\usepackage{subfigure}
\usepackage{booktabs} 
\usepackage{hyperref}

\usepackage{icml2024}
\makeatletter
\@ifpackageloaded{lineno}{\pagewiselinenumbers\nolinenumbers}{}  
\makeatother

\usepackage{amsmath}
\usepackage{amssymb}
\usepackage{mathtools}
\usepackage{amsthm}

\usepackage[capitalize,noabbrev]{cleveref}

\theoremstyle{plain}

\theoremstyle{definition}

\theoremstyle{remark}

\usepackage[textsize=tiny]{todonotes}

\icmltitlerunning{Taylor Series Neural Networks for Time Series Prediction}

\begin{document}

\twocolumn[
\icmltitle{Incorporating Taylor Series and Recursive Structure in Neural Networks for Time Series Prediction}



\icmlsetsymbol{equal}{*}

\begin{center}
Jarrod Mau and Kevin Moon\\
Department of Mathematics and Statistics, Utah State University, Logan, UT, USA
\end{center}

\icmlkeywords{Machine Learning, taylor series, time series,  predictions, neural network}

\vskip 0.3in
]




\begin{abstract}
Time series analysis is relevant in various disciplines such as physics, biology, chemistry, and finance. In this paper, we present a novel neural network architecture that integrates elements from ResNet structures, while introducing the innovative incorporation of the Taylor series framework. This approach demonstrates notable enhancements in test accuracy across many of the baseline datasets investigated. Furthermore, we extend our method to incorporate a recursive step,  which leads to even further improvements in test accuracy. Our findings underscore the potential of our proposed model to significantly advance time series analysis methodologies, offering promising avenues for future research and application.
\end{abstract}

\section{Introduction}
\label{intro}
Time series prediction plays a pivotal role in numerous real-world applications, wielding transformative potential across diverse domains. In the realm of finance, accurate time series forecasting guides investment decisions, risk assessment, and portfolio management \cite{SEZER2020106181}. For weather and climate science, robust predictions enable early warning systems for natural disasters, optimization of energy resources, and proactive agricultural planning \cite{KAREVAN20201}. In healthcare, time series analysis aids in disease outbreak prediction, patient monitoring, and personalized treatment strategies \cite{DBLP:journals/corr/abs-2010-12493, article_2}. Manufacturing industries leverage these techniques to enhance supply chain management, maintenance scheduling, and production optimization \cite{DBLP:journals/corr/abs-2010-12493, article_2}. Moreover, time series forecasting supports digital marketing by anticipating consumer trends and optimizing advertisement placement \cite{armstrong1999forecasting}. Transportation systems rely on it for traffic prediction and route optimization \cite{sabry2007time}. In summary, accurate predictions of time-evolving data empower decision-makers across industries, fostering informed choices, efficient resource allocation, and the ability to proactively address challenges, thus underscoring the critical importance of advancing neural network structures for improved time series analysis \cite{HEWAMALAGE2021388}.

Time series analysis plays a pivotal role in extracting valuable insights from sequential data, uncovering patterns, trends, and underlying structures that drive temporal dynamics \citep{ZHANG2003159, TANG1991}. The ubiquity of time series data across diverse domains, including finance, healthcare, and environmental science, underscores the critical need for accurate and efficient analytical methods \citep{SAGHEER2019203, SEZER2020106181}. Traditional time series models often grapple with the intricate patterns present in real-world datasets, motivating the exploration of innovative approaches \citep{KHASHEI2010479, ZHANG2003159}.

Early forays into time series analysis primarily relied on classical statistical methods, such as autoregressive integrated moving average (ARIMA) models and exponential smoothing techniques \citep{ZHANG2003159, TANG1991}. While effective for certain applications, these methods often struggled to capture the intricacies of non-linear and dynamic temporal patterns \citep{TANG1991, HEWAMALAGE2021388}, prompting the exploration of more sophisticated approaches.

The advent of machine learning marked a paradigm shift in time series analysis, with researchers turning to algorithms capable of learning complex dependencies and patterns from data \citep{SAGHEER2019203, SEZER2020106181}. Support Vector Machines (SVM), Random Forests, and k-Nearest Neighbors (k-NN) emerged as prominent players, showcasing improved predictive capabilities. However, the inherent limitations of these models in handling sequential dependencies and long-range temporal patterns paved the way for the dominance of neural networks.

The rise of deep learning, particularly recurrent neural networks (RNNs) \citep{cho-etal-2014-learning} and Long Short-Term Memory (LSTM) networks \citep{SAGHEER2019203}, revolutionized time series analysis by enabling the modeling of sequential dependencies over extended temporal contexts. These architectures demonstrated remarkable success in various applications, from financial forecasting \citep{SEZER2020106181} to healthcare predictions \citep{DBLP:journals/corr/abs-2010-12493}. Despite their achievements, challenges such as vanishing gradients and difficulty in capturing seasonality persist \citep{HEWAMALAGE2021388}.

Contemporary research in time series analysis grapples with addressing the limitations of existing models, adapting to the heterogeneity of real-world datasets, and improving interpretability \citep{ZHANG2003159, SEZER2020106181}. As datasets grow in complexity, there is a growing need for models that can efficiently navigate through irregularities, non-linearities, and noisy signals inherent in time series data.

Against this backdrop, we introduce a state-of-the-art neural network structure called TaylorNet that is designed to enhance test accuracy in univariate time series analysis. TaylorNet works by amalgamating the strengths of existing architectures and introducing innovative modifications \citep{Mau2023}. By integrating state-of-the-art neural network architectures and introducing novel modifications, TaylorNet seeks to push the boundaries of predictive accuracy, offering a robust solution to the challenges posed by diverse and intricate temporal datasets.

Our contributions are as follows: 1) we present TaylorNet, a novel neural network modification based on a Taylor series formulation of the time series. 2) We compare TaylorNet with other state-of-the-art methods for time series analysis including ResNet \citep{DBLP:journals/corr/abs-2010-12493} and an LSTM \citep{SAGHEER2019203}. 3) We define a recursive version of TaylorNet that shows further improvements in prediction accuracy.





\section{TaylorNet}

Here we define the TaylorNet architecture. Assume we have  $n$ data points in a univariate time series $\{x_1,...,x_n\}$. We break these into sub-sequences of length $d$: $\{(x_j)_{j=0}^{d},...,(x_j)_{j=n-d}^{n}\}.$ All but the last point in each subsequence (i.e. $(x_j)_{j=i}^{i+d-1}$) will be fed into the neural network to output a prediction for the final point $x_{i+d}$.

For time series analysis, a 1-dimensional Convolutional Neural Network (CNN) is often used \cite{7870510}. The output of the 1D CNN can be written as:
\begin{equation}
\hat{x}_{i+d} = \mathcal{N}\left((x_j)_{j=i}^{i+d-1}\right),
\end{equation}
where $\mathcal{N}$ is a feed forward neural network that has a $d-1$ dimensional input layer that consists of the $d-1$ values of the time series prior to the value $x_{i+d}$. The neural network is then trained to predict $x_{i+d}$. 

Alternatively ResNet (Residual Neural Network) can be used for time series analysis \cite{8587554}. ResNet is a groundbreaking neural network architecture that has shown exceptional performance in various computer vision tasks, primarily due to its ability to address the vanishing gradient problem and enable the training of extremely deep networks. While ResNet was initially designed for image analysis, its principles can be extended to time series analysis and prediction with promising outcomes. 

In the context of time series, ResNet's core innovation lies in the introduction of residual or skip connections that enable the network to learn and model the residual information (difference between predicted and actual values) of the input data. This residual learning allows the network to efficiently capture temporal dependencies and patterns present in sequential data, making it especially well-suited for time series forecasting. By incorporating residual connections, ResNet architectures can learn to model the complex relationships within time series data \cite{8587554}.

For time series analysis, the ResNet model prediction can be written as
\begin{equation}
\hat{x}_{i+d} = x_{i+d-1}+\mathcal{N}\left((x_j)_{j=i}^{i+d-1}\right), \label{eq:ResNet}
\end{equation}
where again $\mathcal{N}$ is a feed forward neural network that takes as input the previous $d-1$ time series values. 
In the ResNet model, the output is $x_{i+d-1}$ plus the output of the neural network $\mathcal{N}$. In this case, the neural network models the residual, in other words, the change for the given timestep.

\begin{figure}[H]
    \centering
    \includegraphics[width=1\columnwidth]{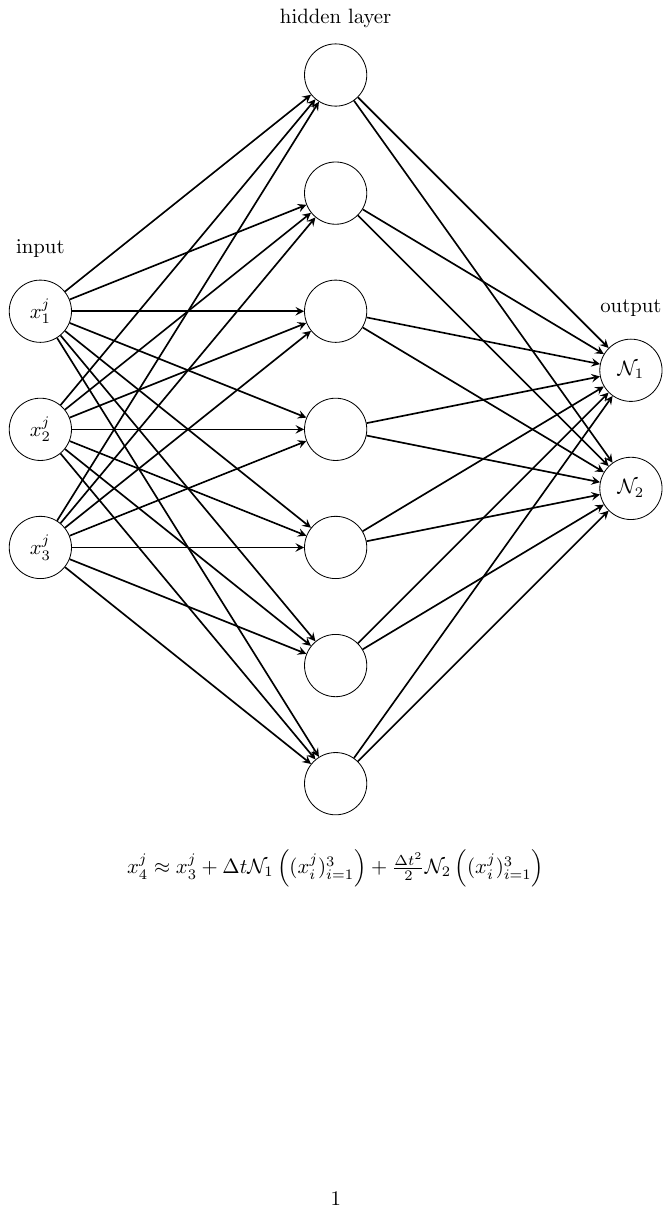}
    \caption{Example of a TaylorNet2 architecture with an input sequence length of 3 and a single hidden layer. The outputs $\mathcal{N}_1$ and $\mathcal{N}_2$ approximate the derivatives $\frac{dx}{dt}$ and $\frac{d^2x}{dt^2}$, respectively, of the underlying dynamics of the time series. }
    \label{fig:taylor2}
\end{figure}

Using the lens of Taylor series approximations we can think of the neural network $\mathcal{N}$ as an approximation of the derivative $\frac{dx}{dt}$. Thus the ResNet model is very similar to  a first order Taylor series approximation of the time series, which can be written as:
\begin{equation}
\hat{x}_{i+d} = x_{i+d-1}+\Delta t\mathcal{N}\left((x_j)_{j=i}^{i+d-1}\right)
\label{taylor1equation}
\end{equation}
Note the addition of the $\Delta t$ factor, which will come into play later. We can thus extend this approach to higher-order Taylor series approximations. We first propose TaylorNet2, which uses a second-order expansion as follows: 
\begin{equation}
    \hat{x}_{i+d} = x_{i+d-1} + \Delta t \mathcal{N}_1\left((x_j)_{j=i}^{i+d-1}\right) + \frac{\Delta t^2}{2} \mathcal{N}_2\left((x_j)_{j=i}^{i+d-1}\right)
\label{taylor2equation}
\end{equation}
where $\mathcal{N}_1$ and $\mathcal{N}_2$ are the outputs of a feedforward neural network that takes as input the previous $d-1$ time series values. Figure~\ref{fig:taylor2} shows an example of such a network with a single hidden layer when $d=4$.

\begin{figure}[t!]
    \centering
    \includegraphics[width=1\columnwidth]{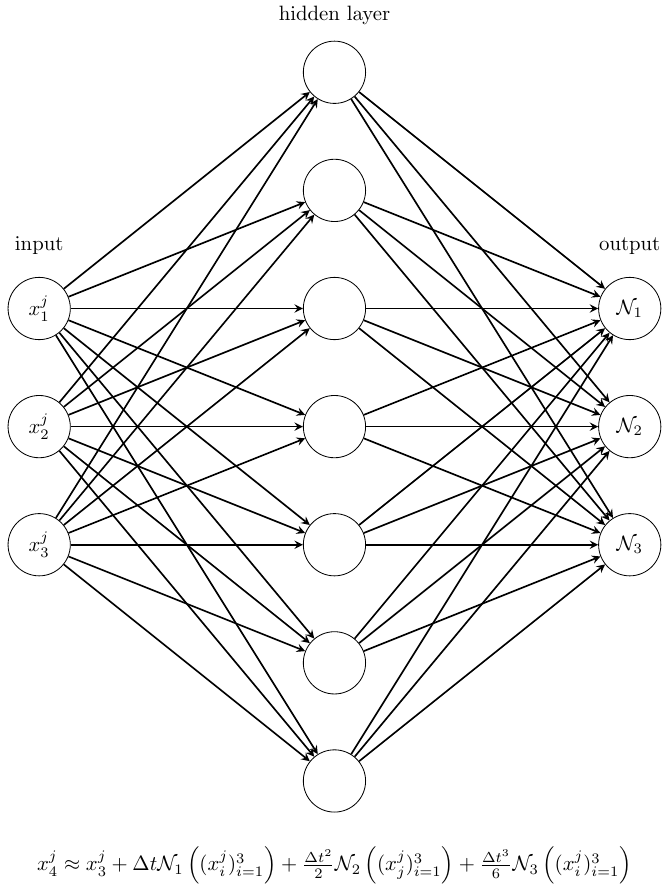}
    \caption{Example of a TaylorNet3 archietcture with an input sequence length of 3 and a single hidden layer. The outputs $\mathcal{N}_1$, $\mathcal{N}_2$, and $\mathcal{N}_3$ approximate the derivatives $\frac{dx}{dt}$, $\frac{d^x}{dt^2}$, and $\frac{d^3x}{dt^3}$, respectively, of the underlying dynamics of the time series.}
    \label{fig:taylor3}
\end{figure}

TaylorNet3 is defined similarly:
\begin{align}
    \hat{x}_{i+d} =& x_{i+d-1} + \Delta t \mathcal{N}_1\left((x_j)_{j=i}^{i+d-1}\right)\\  &+ \frac{\Delta t^2}{2} \mathcal{N}_2\left((x_j)_{j=i}^{i+d-1}\right)  \nonumber +  \frac{\Delta t^3}{6} \mathcal{N}_3\left((x_j)_{j=i}^{i+d-1}\right),
\label{taylor3equation}
\end{align}
where $\mathcal{N}_1$, $\mathcal{N}_2$, and $\mathcal{N}_3$ are again the outputs of a feedforward network that takes as input the previous $d-1$ time series values. Figure~\ref{fig:taylor3} shows an example of such a network with a single hidden layer when $d=4$.


In both frameworks, the neural network outputs predictions for  $\frac{dx}{dt}$ and$\frac{d^2x}{dt^2}$. Additionally, TaylorNet3 outputs a prediction for $\frac{d^3x}{dt^3}$. The authors in \cite{Mau2023} provide some empirical evidence that the TaylorNet framework will be successful. In \cite{Mau2023}, the authors used similar structures to successfully learn higher order information with time lagged data. This approach was used specifically to learn the physics of ordinary differential systems of equations, but the same principle applies to time series data. While it is difficult to show theoretically that the neural network will approximate the higher order information given it is a black box function, the authors demonstrated empirically that the neural network was able to learn the higher order information when using toy examples where the underlying dynamics are known \cite{Mau2023}.

\subsection{Recursive TaylorNet}
\label{sub:recursive}

We can generalize the standard TaylorNet approach described previously by adding recursion. In the standard TaylorNet framework, $\Delta t$ is chosen to be 1 to match the time difference between data points in the time series. To add recursion, we can instead  take several smaller steps to get to the next value in the time series. This is motivated by basic numerical methods of solving Ordinary Differential Equations \cite{doi:10.1137/0709052}. 

To better illustrate this idea consider an example where we use a sequence length of 3 for an input into our neural network and use 2 steps to predict the next value in our time series. To do this, we start with the time series input data $x_1, x_2, x_3$, which is used to predict $x_4$. Consider the structure in Eq.~\eqref{taylor1equation}. To add recursion, we first halve the time step and modify the neural network to have three outputs that correspond with the first order information for each value in our input sequence:
\begin{equation}
    \mathcal{N}(x_1, x_2, x_3) = (v_1, v_2, v_3).
\end{equation}

We can then project the initial state 1/2 time step into the future using the first order information outputted by the feed forward neural network: 
\begin{equation}
    (\hat{x}_{1.5}, \hat{x}_{2.5}, \hat{x}_{3.5}) = (x_1, x_2, x_3) + \frac{\Delta t}{2}(v_1, v_2, v_3).
\end{equation}
We then plug these values back into the neural network to get the corresponding first order information:
\begin{equation}
    \mathcal{N}(\hat{x}_{1.5}, \hat{x}_{2.5}, \hat{x}_{3.5}) = (v_{1.5}, v_{2.5}, v_{3.5}).
\end{equation}
We then project the state again 1/2 time step into the future to get the prediction for $x_4$: 
\begin{equation}
    (\hat{x}_2, \hat{x}_3, \hat{x}_4) = (\hat{x}_{1.5}, \hat{x}_{2.5}, \hat{x}_{3.5}) + \frac{\Delta t}{2}(v_{1.5}, v_{2.5}, v_{3.5}).
\end{equation}

We can generalize this setting to arbitrary sequence lengths and arbitrary numbers of recursion steps. The general equation is illustrated below with an input sequence of $d-1$ and a step size of $1/m$:

\begin{equation}
\left(\hat{x}_{j_{k+1}}\right)_{j=i}^{i+d-1} = \left(\hat{x}_{j_k}\right)_{j=i}^{i+d-1}+\frac{\Delta t}{m}\mathcal{N}\left(\left(\hat{x}_{j_k}\right)_{j=i}^{i+d-1}\right),
\end{equation}

where 

\begin{equation}
    \hat{x}_{j_k}=\hat{x}_{j+\frac{k}{m}},
\end{equation}

and

\begin{equation}
 \left(\hat{x}_{j_0}\right)_{j=i}^{i+d-1}=\left({x}_{j}\right)_{j=i}^{i+d-1}
\end{equation}

is the initial input that starts the recursion process.

This leads us  to our proposed Recursive TaylorNet2  and Recursive TaylorNet3: 

\begin{equation}
\begin{aligned}
    (\hat{x}_{j_{k+1}})_{j=i}^{i+d-1} &= (\hat{x}_{j_{k}})_{j=i}^{i+d-1} + \\ 
    &\Delta t (\mathcal{N}_1^k)_{k=1}^d + \frac{\Delta t^2}{2} (\mathcal{N}_2^k)_{k=1}^d,
\end{aligned}
\end{equation}
where 
\begin{equation}
\mathcal{N}((x_{j_{k}})_{j=i}^{i+d-1})= \left((\mathcal{N}_1^k)_{k=1}^d , (\mathcal{N}_2^k)_{k=1}^d\right)
\end{equation}
are the equations for TaylorNet2. For TaylorNet3, we have:

\begin{equation}
\begin{aligned}
    (\hat{x}_{j_{k+1}})_{j=i}^{i+d-1}  = (\hat{x}_{j_{k}})_{j=i}^{i+d-1} + \Delta t (\mathcal{N}^k_1)_{k=1}^d \\  + \frac{\Delta t^2}{2} (\mathcal{N}^k_2)_{k=1}^d  +  \frac{\Delta t^3}{6} (\mathcal{N}^k_3)_{k=1}^d,
\end{aligned}
\end{equation}
where
\begin{equation}
\mathcal{N}((x_j)_{j=i}^{i+d-1})= \left((\mathcal{N}^k_1)_{k=1}^d, (\mathcal{N}^k_2)_{k=1}^d, (\mathcal{N}^k_3)_{k=1}^d \right).
\end{equation}

\section{Results}
\subsection{TaylorNet Results}

\begin{table}
\caption{Descriptions of datasets}
\centering
\begin{tabular}{ |p{0.12\textwidth} |p{0.2\textwidth} | p{0.06\textwidth}|}
\hline
\textbf{Data} & \textbf{Description} & \textbf{Sample Size} \\
\hline
EuStock \cite{R-dataset-EuStock}& Daily closing prices of the major European stock indice DAX  & 1860\\
\hline
UKgas \cite{R-dataset-UKgas}& Quarterly UK gas consumption from 1960Q1 to 1986Q4  & 108\\
\hline
austres \cite{R-dataset-austres}& Numbers of Australian residents quarterly from 1971 to 1994  & 89\\
\hline
discoveries \cite{R-dataset-discoveries}& The numbers of “great” discoveries in each year from 1860 to 1959  & 100\\
\hline
treering \cite{R-dataset-treering}& Yearly Treering Data, -6000–1979  & 7980\\
\hline
WWWusage \cite{R-dataset-WWWusage} & Users connected to the Internet through a server by minute.  & 100\\
\hline
BJsales \cite{R-dataset-BJsales} & Box and Jenkins Sales  & 150\\
\hline
LakeHuron \cite{R-dataset-LakeHuron}& Annual measurements of the level, in feet, of Lake Huron 1875–1972  & 98 \\
\hline
Seatbelts \cite{R-dataset-Seatbelts}  & Monthly totals of car drivers killed or seriously injured 1969-1984 & 192\\
\hline
Energy \cite{Household-Power-Consumption} & Electricity consumption by minute 2006-2010  & 1000\\
\hline
airline \cite{R-dataset-AirPassengers} & Monthly airline passengers 1949-1960  & 144\\
\hline
sunspots \cite{Sunspots} & Monthly sunspot numbers from 1749 to 2019  & 3291\\
\hline
Nile \cite{Nile} & Annual depth measurements of the River Nile from 1871 to 1970  & 100\\
\hline
Lynx  \cite{Lynx} & Annual counts of lynx trapped in Canada from 1821 to 1934 & 114\\
\hline
co2 \cite{CO2-PPM} & Atmospheric Carbon Dioxide  & 727\\
\hline
temperature \cite{TemperatureData} & Daily temperature from 1880 to 2014  & 3650\\
\hline
milk \cite{MilkProductionData} & Monthly milk production: pounds per cow. 1962 - 1975  & 168\\
\hline
pressure \cite{AtmosphericPressureData} & Hourly pressure data  & 5000\\
\hline
\end{tabular}
\label{tab:descriptions}

\end{table}

\begin{table*}
\caption{Best test error for each network for each data set over all learning rates, initializations, and sequence lengths. The results of the best performing method for each dataset are highlighted. The average rank and median percent deviation from the best performer for each method are given.  TaylorNet2 outperforms all other methods based on the average rank, and ties with ResNet based on the median percent deviation.}
\centering
\begin{tabular}{ |c |c |c |c|c |c|}
\hline
 &  \multicolumn{3}{c|}{Baseline Models} & \multicolumn{2}{c|}{Novel Models}\\
\hline
 & CNN  & ResNet  & LSTM  & Taylor 2 & Taylor 3\\
\hline
EuStock & 0.001158 & 0.000146 & 0.000303 & \cellcolor{green!25}0.000146 & 0.000146 \\
\hline
UKgas & 0.034143 & \cellcolor{green!25}0.001672 & 0.013483 & 0.001729 & 0.001691 \\
\hline
austres  & 0.007553 & 0.000010 & 0.002904 & \cellcolor{green!25}0.000010 & 0.000045 \\
\hline
discoveries  & \cellcolor{green!25}0.011287 & 0.013375 & 0.013434 & 0.012841 & 0.012716 \\
\hline
treering  & 0.020445 & 0.020069 & 0.020395 & \cellcolor{green!25}0.020020 & 0.020023 \\
\hline
WWWusage  & 0.011866 & 0.000647 & 0.002065 & \cellcolor{green!25}0.000628 & 0.000857 \\
\hline
BJsales  & 0.000624 & 0.000212 & 0.001952 & \cellcolor{green!25}0.000184 & 0.000222 \\
\hline
LakeHuron  & 0.036395 & 0.015338 & 0.025124 & 0.014586 & \cellcolor{green!25}0.013411 \\
\hline
Seatbelts  & 0.020450 & \cellcolor{green!25}0.011825 & 0.022714 & 0.011933 & 0.012705 \\
\hline
Energy  & 0.014987 & 0.010365 & 0.012572 &  0.009729 & \cellcolor{green!25}0.009551 \\
\hline
airline  & 0.016180 & 0.000847 & 0.013128 & \cellcolor{green!25}0.000836 & 0.000846 \\
\hline
sunspots  & 0.003514 & 0.003553 & 0.003539 &\cellcolor{green!25}0.003505 & 0.003612 \\
\hline
Nile  & \cellcolor{green!25}0.015781 & 0.016015 & 0.015994 & 0.016078 & 0.016063 \\
\hline
Lynx  & 0.010403 & 0.003729 & 0.012896  & 0.003616 & \cellcolor{green!25}0.003512 \\
\hline
co2  & 0.001371 & \cellcolor{green!25}0.000049 & 0.000228 & 0.000060 & 0.000064  \\
\hline
temperature  & 0.007996 & 0.007807 & \cellcolor{green!25}0.007744 & 0.007823 & 0.007773 \\
\hline
milk  & 0.011843 & \cellcolor{green!25}0.000437 & 0.016577 & 0.000682 & 0.000589 \\
\hline
pressure  & \cellcolor{green!25}0.010234 & 0.011001 & 0.011019 & 0.010989 & 0.011121 \\
\hline
Average Rank & 3.94 & 2.44 & 3.94 & 2.06 & 2.61 \\
\hline
Median Percent Deviation & 183.8 & 1.4 & 99.8 & 1.4 & 2.4\\
\hline
\end{tabular}
\label{tab:taylor_combined}

\end{table*}

Here we compare our TaylorNet architectures to three baseline neural networks. These models include a standard 1D CNN, a modified 1D CNN with a ResNet framework (as in Eq.~\eqref{eq:ResNet}), and an LSTM.  We tested the models  on 18 data sets that are univariate time series with sample sizes ranging from 89 to 5000. We chose the time series data  to cover a wide range of dynamics in finance, biology, weather, logistics, and physics. See Table~\ref{tab:descriptions} for details on all of the datasets. Because we are focusing on relatively small sample sizes, we did not compare our approach to a transformer model, which typically require much larger sample sizes. 

We used learning rates of 0.1, 0.01, and 0.001 and sequence lengths of 3, 5, 7, 9, 11, and 13. We used  3 initializations with early stopping. Each neural network was a standard feed forward neural network with 128 nodes in one hidden layer, a sigmoid activation function on the hidden layer, and a linear activation in the output layer. The reason we used a shallow network was to strip away the complexities in the models, giving us confidence that any differences in performance between the baseline methods and the TaylorNet architectures are likely due to our proposed modifications.  

The results of the experiments are given in Table~\ref{tab:taylor_combined}. From these results, we see that the ResNet structure does the best in terms of test accuracy on average compared to the other two baseline models. However, the TaylorNet2 architecture performs the best overall, with the lowest average rank. TaylorNet2 and ResNet perform similarly using the median percent deviation from the best performer. TaylorNet3 also performs well, although not as well as TaylorNet2 and ResNet. This may be due to the extra complexity of TaylorNet3 as well as the additional assumption of the underlying dynamics being 3rd order differentiable, which may be overly restrictive.  

While these results suggest that there is untapped potential in incorporating second and third order derivative information in the neural network structure, there may be diminishing returns as evidenced by the underperformance of TaylorNet3 relative to TaylorNet2.  Incorporating even higher order information could improve accuracy theoretically, but it would come at the cost of increasing the complexity of the model as well as imposing the assumption that the underlying dynamics are higher order differentiable. Assuming higher order differentiability would be fair  for datasets measuring sunspot activity (i.e. the sunspots dataset), which is a natural phenomena. Thus incorporating second and third order information into the structure of a neural network seems to help in these cases. However, assuming higher order differentiability is less likely to be fair for discoveries which stem form discrete discoveries for a given year as in the discoveries dataset. Discoveries for a given year could be first order differentiable but are unlikely to be reliably third- or fourth-order differentiable. 

\begin{table*}
\caption{Best test error for each network for each data set over all learning rates, initializations, and sequence lengths. The results of the best performing method for each dataset are highlighted. The average rank and median percent deviation from the best performer are given. Recursive TaylorNet2 (denoted R Taylor 2) outperforms all other methods on average based on these metrics, with Recursive TaylorNet 3 (R Taylor 3) and Recursive ResNet also performing well. }
\centering
\begin{tabular}{ |c |c |c |c|c |c|c|}
\hline
 &  \multicolumn{3}{c|}{Baseline Models} & \multicolumn{3}{c|}{Novel Models}\\
\hline
 & CNN  & ResNet  & LSTM   & R ResNet & R Taylor 2 & R Taylor 3 \\
\hline
EuStock & 0.001158 & 0.000146 & 0.000303 & 0.000145 & \cellcolor{green!25}0.000145 & 0.000146 \\
\hline
UKgas & 0.034143 & 0.001672 & 0.013483 & 0.001611 & \cellcolor{green!25}0.001465 & 0.001535 \\
\hline
austres  & 0.007553 & 0.000010 & 0.002904 & 0.000020 & \cellcolor{green!25}0.000009 & 0.000009 \\
\hline
discoveries  & \cellcolor{green!25}0.011287 & 0.013375 & 0.013434 & 0.012942 & 0.012518 & 0.012383 \\
\hline
treering  & 0.020445 & 0.020069 & 0.020395 & \cellcolor{green!25}0.020019 & 0.020173 & 0.020124 \\
\hline
WWWusage  & 0.011866 & 0.000647 \cellcolor{green!25} & 0.002065 & 0.000698 & 0.000677 & 0.000688 \\
\hline
BJsales  & 0.000624 & 0.000212 & 0.001952 &  0.000188 & \cellcolor{green!25}0.000172 & 0.000181 \\
\hline
LakeHuron  & 0.036395 & 0.015338 & 0.025124 & \cellcolor{green!25}0.014404 & 0.014772 & 0.014807 \\
\hline
Seatbelts  & 0.020450 & 0.011825 & 0.022714 & 0.011910 & \cellcolor{green!25}0.010594 & 0.011873 \\
\hline
Energy  & 0.014987 & 0.010365 & 0.012572& 0.009799 & \cellcolor{green!25}0.009429 & 0.009526 \\
\hline
airline  & 0.016180 & \cellcolor{green!25}0.000847 & 0.013128 &  0.001009 & 0.001261 & 0.001565 \\
\hline
sunspots  & 0.003514 & 0.003553 & 0.003539 & 0.003508 & \cellcolor{green!25}0.003491 & 0.003580 \\
\hline
Nile  & 0.015781 & 0.016015 & 0.015994 &\cellcolor{green!25}0.014724 & 0.014969 & 0.015267 \\
\hline
Lynx  & 0.010403 & 0.003729 & 0.012896  & 0.003240 & 0.003055 & \cellcolor{green!25}0.002929 \\
\hline
co2  & 0.001371 & 0.000049 & 0.000228 & 0.000051 & \cellcolor{green!25}0.000047 & 0.000071 \\
\hline
temperature  & 0.007996 & 0.007807 & \cellcolor{green!25}0.007744 & 0.007802 & 0.007808 & 0.007827 \\
\hline
milk  & 0.011843 & \cellcolor{green!25}0.000437 & 0.016577 & 0.001300 & 0.000639 & 0.000648 \\
\hline
pressure  & \cellcolor{green!25}0.010234 & 0.011001 & 0.011019 & 0.010235 & 0.010434 & 0.010886 \\
\hline
Average Rank & 4.94 & 3.28 & 5.06 & 2.67 & 1.94 & 3.11 \\
\hline
Median Percent Deviation & 203.9 & 7.0 & 111.7 & 8.2 & 0.8 & 4.2 \\
\hline
\end{tabular}
\label{tab:taylor_r_combined}
\end{table*}

\subsection{Recursive TaylorNet Results}
Here we explore the performance of Recursive TaylorNet, which was introduced in Section~\ref{sub:recursive}.
 We know from numerical analysis that adding a recursive step will improve accuracy in the projections of dynamical systems but, just as before, this improvement requires a continuous and differentiable system. Since this isn't always true, we do not expect Recursive TaylorNet to work well  on all data sets. 
 
 We denote the recursive versions of our previous neural networks as R ResNet, R Taylor 2, and R Taylor 3. They each have ascending levels of potential in terms of theoretical test accuracy \cite{Mau2023} and ascending levels of complexity and assumptions which could impact the results negatively. We used the same setup as before: 18 data sets with learning rates 0.1, 0.01, and 0.001 with sequence lengths 3, 5, 7, 9, 11, and 13 over 3 initializations. This method requires another hyperperameter, substeps: the number of recursive steps required to arrive at $x_{i+d}$. We considered 2, 3, 4, 5, 6, 7, and 8 recursive steps.

The results are given in Table~\ref{tab:taylor_r_combined}. These results show that the proposed recursive methods outperformed the traditional methods in two thirds of the data sets we looked into, which is better than the proposed methods without the recursive step. This  discovery  attests to the potential value of this recursive framework. We see that the recursive methods do well on almost all of the same datasets as the standard TaylorNet architectures, which makes sense given they require the same assumptions of the underlying dynamics. 

We see also that R Taylor 2 performs the best overall in terms of the average rank and the median percent deviation from the best performer. Thus R Taylor 2 is the clear winner in time series forecasting across this diverse set of data. This suggest that R Taylor 2 may balance the bias-variance tradeoff well while not requiring as many assumptions as R Taylor 3 (i.e. third order differentiability of the underlying dynamics).

\section{Discussion}
Our results suggest that any model that does time series predictions is likely to improve by incorporating the recursive Taylor series structure to some degree. The model structure doesn't need to change significantly. Furthermore any model, relating to time series or otherwise, that already has a ResNet structure could benefit from looking at it from a Taylor series lens and incorporating higher order terms or a recursive step. This is an exciting new avenue of research that can span across multiple domains of machine learning.

\section{Conclusion}
In conclusion, our study presents an innovative approach to time series analysis through the development of the TaylorNet architecture. We have demonstrated the efficacy of our proposed model in improving test accuracy across various univariate time series datasets, showcasing its potential to outperform traditional methods and even state-of-the-art neural network architectures such as ResNet and LSTM.

By incorporating elements from ResNet structures and integrating the Taylor series framework, our TaylorNet architecture introduces a novel paradigm for modeling temporal dependencies and patterns in sequential data. Through empirical evaluations on a diverse range of datasets, we have shown that TaylorNet, particularly the TaylorNet2 variant, offers notable enhancements in predictive accuracy compared to baseline models.

Furthermore, we extended our proposal to include a recursive step, allowing for multiple smaller steps to predict future values in the time series. This recursive TaylorNet approach demonstrated even further improvements in test accuracy, outperforming both traditional and proposed models without the recursive step in a significant portion of the datasets examined.

Our findings underscore the potential of TaylorNet and its recursive variants to advance the field of time series analysis, offering a promising avenue for researchers and practitioners seeking heightened accuracy in forecasting temporal data. Moreover, the flexibility of our architecture, coupled with its ability to capture higher-order derivative information, opens up exciting opportunities for exploring more complex temporal dynamics and real-world applications across various domains.

As we continue to refine and expand upon our proposed framework, future research directions may include investigating the scalability of TaylorNet to larger and more diverse datasets, exploring additional variations and extensions of the architecture, and delving deeper into the theoretical underpinnings of its performance. Ultimately, TaylorNet represents a significant step forward in the quest for more accurate and reliable time series forecasting methodologies, with implications spanning multiple fields and disciplines.

\section{Impact Statement}

This paper presents work whose goal is to advance the field of Machine Learning. There are many potential societal consequences of our work, none which we feel must be specifically highlighted here.


\begin{thebibliography}{34}
\providecommand{\natexlab}[1]{#1}
\providecommand{\url}[1]{\texttt{#1}}
\expandafter\ifx\csname urlstyle\endcsname\relax
  \providecommand{\doi}[1]{doi: #1}\else
  \providecommand{\doi}{doi: \begingroup \urlstyle{rm}\Url}\fi

\bibitem[Armstrong \& Brodie(1999)Armstrong and Brodie]{armstrong1999forecasting}
Armstrong, J.~S. and Brodie, R.
\newblock Forecasting for marketing.
\newblock pp.\  92--119, 1999.
\newblock URL \url{https://ssrn.com/abstract=662622}.

\bibitem[Chaudhari(2020)]{MilkProductionData}
Chaudhari, C.
\newblock Prediction of monthly milk production time series.
\newblock \url{https://www.kaggle.com/code/chandraveshchaudhari}, 2020.

\bibitem[Cho et~al.(2014)Cho, van Merri{\"e}nboer, Gulcehre, Bahdanau, Bougares, Schwenk, and Bengio]{cho-etal-2014-learning}
Cho, K., van Merri{\"e}nboer, B., Gulcehre, C., Bahdanau, D., Bougares, F., Schwenk, H., and Bengio, Y.
\newblock Learning phrase representations using {RNN} encoder{--}decoder for statistical machine translation.
\newblock In Moschitti, A., Pang, B., and Daelemans, W. (eds.), \emph{Proceedings of the 2014 Conference on Empirical Methods in Natural Language Processing ({EMNLP})}, pp.\  1724--1734, Doha, Qatar, October 2014. Association for Computational Linguistics.
\newblock \doi{10.3115/v1/D14-1179}.
\newblock URL \url{https://aclanthology.org/D14-1179}.

\bibitem[Choi et~al.(2018)Choi, Ryu, and Kim]{8587554}
Choi, H., Ryu, S., and Kim, H.
\newblock Short-term load forecasting based on resnet and lstm.
\newblock In \emph{2018 IEEE International Conference on Communications, Control, and Computing Technologies for Smart Grids (SmartGridComm)}, pp.\  1--6, 2018.
\newblock \doi{10.1109/SmartGridComm.2018.8587554}.

\bibitem[Guhr(2023)]{TemperatureData}
Guhr, O.
\newblock Daily minimum temperatures data.
\newblock \url{https://github.com/oliverguhr/transformer-time-series-prediction/blob/master/daily-min-temperatures.csv}, 2023.

\bibitem[Hewamalage et~al.(2021)Hewamalage, Bergmeir, and Bandara]{HEWAMALAGE2021388}
Hewamalage, H., Bergmeir, C., and Bandara, K.
\newblock Recurrent neural networks for time series forecasting: Current status and future directions.
\newblock \emph{International Journal of Forecasting}, 37\penalty0 (1):\penalty0 388--427, 2021.
\newblock ISSN 0169-2070.
\newblock \doi{https://doi.org/10.1016/j.ijforecast.2020.06.008}.
\newblock URL \url{https://www.sciencedirect.com/science/article/pii/S0169207020300996}.

\bibitem[Hull et~al.(1972)Hull, Enright, Fellen, and Sedgwick]{doi:10.1137/0709052}
Hull, T.~E., Enright, W.~H., Fellen, B.~M., and Sedgwick, A.~E.
\newblock Comparing numerical methods for ordinary differential equations.
\newblock \emph{SIAM Journal on Numerical Analysis}, 9\penalty0 (4):\penalty0 603--637, 1972.
\newblock \doi{10.1137/0709052}.
\newblock URL \url{https://doi.org/10.1137/0709052}.

\bibitem[Iverson(2023)]{Household-Power-Consumption}
Iverson, M.
\newblock Household power consumption.
\newblock \url{https://github.com/mkivenson/Household-Power-Consumption}, 2023.

\bibitem[Karevan \& Suykens(2020)Karevan and Suykens]{KAREVAN20201}
Karevan, Z. and Suykens, J.~A.
\newblock Transductive lstm for time-series prediction: An application to weather forecasting.
\newblock \emph{Neural Networks}, 125:\penalty0 1--9, 2020.
\newblock ISSN 0893-6080.
\newblock \doi{https://doi.org/10.1016/j.neunet.2019.12.030}.
\newblock URL \url{https://www.sciencedirect.com/science/article/pii/S0893608020300010}.

\bibitem[Khashei \& Bijari(2010)Khashei and Bijari]{KHASHEI2010479}
Khashei, M. and Bijari, M.
\newblock An artificial neural network (p,d,q) model for timeseries forecasting.
\newblock \emph{Expert Systems with Applications}, 37\penalty0 (1):\penalty0 479--489, 2010.
\newblock ISSN 0957-4174.
\newblock \doi{https://doi.org/10.1016/j.eswa.2009.05.044}.
\newblock URL \url{https://www.sciencedirect.com/science/article/pii/S0957417409004850}.

\bibitem[Mau \& Zhao(2023)Mau and Zhao]{Mau2023}
Mau, J. and Zhao, J.
\newblock Discovery of governing equations with recursive deep neural networks.
\newblock \emph{Communications on Applied Mathematics and Computation}, 2023.
\newblock ISSN 2661-8893.
\newblock \doi{10.1007/s42967-023-00270-0}.
\newblock URL \url{https://doi.org/10.1007/s42967-023-00270-0}.

\bibitem[{R Core Team}(2022{\natexlab{a}})]{Lynx}
{R Core Team}.
\newblock Canadian lynx trappings, 2022{\natexlab{a}}.
\newblock \url{https://stat.ethz.ch/R-manual/R-devel/library/datasets/html/lynx.html}.

\bibitem[{R Core Team}(2022{\natexlab{b}})]{Nile}
{R Core Team}.
\newblock Nile river flow data, 2022{\natexlab{b}}.
\newblock \url{https://stat.ethz.ch/R-manual/R-devel/library/datasets/html/Nile.html}.

\bibitem[{R Core Team}(2022{\natexlab{c}})]{R-dataset-AirPassengers}
{R Core Team}.
\newblock Airpassengers, 2022{\natexlab{c}}.
\newblock URL \url{https://stat.ethz.ch/R-manual/R-devel/library/datasets/html/AirPassengers.html}.
\newblock {R} package version 4.1.2.

\bibitem[{R Core Team}(2022{\natexlab{d}})]{R-dataset-BJsales}
{R Core Team}.
\newblock \emph{BJsales}, 2022{\natexlab{d}}.
\newblock URL \url{https://stat.ethz.ch/R-manual/R-devel/library/datasets/html/BJsales.html}.
\newblock {R} package version 4.1.2.

\bibitem[{R Core Team}(2022{\natexlab{e}})]{R-dataset-EuStock}
{R Core Team}.
\newblock \emph{EuStock}, 2022{\natexlab{e}}.
\newblock URL \url{https://stat.ethz.ch/R-manual/R-devel/library/datasets/html/EuStock.html}.
\newblock {R} package version 4.1.2.

\bibitem[{R Core Team}(2022{\natexlab{f}})]{R-dataset-LakeHuron}
{R Core Team}.
\newblock \emph{LakeHuron}, 2022{\natexlab{f}}.
\newblock URL \url{https://stat.ethz.ch/R-manual/R-devel/library/datasets/html/LakeHuron.html}.
\newblock {R} package version 4.1.2.

\bibitem[{R Core Team}(2022{\natexlab{g}})]{R-dataset-Seatbelts}
{R Core Team}.
\newblock \emph{Seatbelts}, 2022{\natexlab{g}}.
\newblock URL \url{https://stat.ethz.ch/R-manual/R-devel/library/datasets/html/Seatbelts.html}.
\newblock {R} package version 4.1.2.

\bibitem[{R Core Team}(2022{\natexlab{h}})]{R-dataset-UKgas}
{R Core Team}.
\newblock \emph{UKgas}, 2022{\natexlab{h}}.
\newblock URL \url{https://stat.ethz.ch/R-manual/R-devel/library/datasets/html/UKgas.html}.
\newblock {R} package version 4.1.2.

\bibitem[{R Core Team}(2022{\natexlab{i}})]{R-dataset-WWWusage}
{R Core Team}.
\newblock \emph{WWWusage}, 2022{\natexlab{i}}.
\newblock URL \url{https://stat.ethz.ch/R-manual/R-devel/library/datasets/html/WWWusage.html}.
\newblock {R} package version 4.1.2.

\bibitem[{R Core Team}(2022{\natexlab{j}})]{R-dataset-austres}
{R Core Team}.
\newblock \emph{austres}, 2022{\natexlab{j}}.
\newblock URL \url{https://stat.ethz.ch/R-manual/R-devel/library/datasets/html/austres.html}.
\newblock {R} package version 4.1.2.

\bibitem[{R Core Team}(2022{\natexlab{k}})]{R-dataset-discoveries}
{R Core Team}.
\newblock \emph{discoveries}, 2022{\natexlab{k}}.
\newblock URL \url{https://stat.ethz.ch/R-manual/R-devel/library/datasets/html/discoveries.html}.
\newblock {R} package version 4.1.2.

\bibitem[{R Core Team}(2022{\natexlab{l}})]{R-dataset-treering}
{R Core Team}.
\newblock \emph{treering}, 2022{\natexlab{l}}.
\newblock URL \url{https://stat.ethz.ch/R-manual/R-devel/library/datasets/html/treering.html}.
\newblock {R} package version 4.1.2.

\bibitem[{R Core Team}(2022{\natexlab{m}})]{Sunspots}
{R Core Team}.
\newblock Monthly sunspot data, from 1749 to "1983", 2022{\natexlab{m}}.
\newblock \url{https://stat.ethz.ch/R-manual/R-devel/library/datasets/html/sunspot.month.html}.

\bibitem[Sabry et~al.(2007)Sabry, Abd-El-Latif, Yousef, and Badra]{sabry2007time}
Sabry, M., Abd-El-Latif, H., Yousef, S., and Badra, N.
\newblock A time-series forecasting of average daily traffic volume.
\newblock \emph{Australian Journal of Basic and Applied Sciences}, 1\penalty0 (4):\penalty0 386--394, 2007.
\newblock ISSN 1991-8178.
\newblock Corresponding Author: N. Badra, Department of Physics and Engineering Mathematics, Faculty of Engineering, Ain Shams University, Cairo, Egypt.

\bibitem[Sagheer \& Kotb(2019)Sagheer and Kotb]{SAGHEER2019203}
Sagheer, A. and Kotb, M.
\newblock Time series forecasting of petroleum production using deep lstm recurrent networks.
\newblock \emph{Neurocomputing}, 323:\penalty0 203--213, 2019.
\newblock ISSN 0925-2312.
\newblock \doi{https://doi.org/10.1016/j.neucom.2018.09.082}.
\newblock URL \url{https://www.sciencedirect.com/science/article/pii/S0925231218311639}.

\bibitem[Sezer et~al.(2020)Sezer, Gudelek, and Ozbayoglu]{SEZER2020106181}
Sezer, O.~B., Gudelek, M.~U., and Ozbayoglu, A.~M.
\newblock Financial time series forecasting with deep learning : A systematic literature review: 2005–2019.
\newblock \emph{Applied Soft Computing}, 90:\penalty0 106181, 2020.
\newblock ISSN 1568-4946.
\newblock \doi{https://doi.org/10.1016/j.asoc.2020.106181}.
\newblock URL \url{https://www.sciencedirect.com/science/article/pii/S1568494620301216}.

\bibitem[Sun et~al.(2020)Sun, Hong, Song, and Li]{DBLP:journals/corr/abs-2010-12493}
Sun, C., Hong, S., Song, M., and Li, H.
\newblock A review of deep learning methods for irregularly sampled medical time series data.
\newblock \emph{CoRR}, abs/2010.12493, 2020.
\newblock URL \url{https://arxiv.org/abs/2010.12493}.

\bibitem[Tang et~al.(1991)Tang, de~Almeida, and Fishwick]{TANG1991}
Tang, Z., de~Almeida, C., and Fishwick, P.~A.
\newblock Time series forecasting using neural networks vs. box-jenkins methodology.
\newblock \emph{SIMULATION}, 57\penalty0 (5):\penalty0 303--310, 1991.
\newblock \doi{10.1177/003754979105700508}.

\bibitem[Theodor et~al.(2019)Theodor, Alexandru, and Ianculescu]{article_2}
Theodor, P., Alexandru, A., and Ianculescu, M.
\newblock Assessing and forecasting of epidemiological data using time series analysis.
\newblock 4:\penalty0 37--42, 09 2019.

\bibitem[Tiwari(2021)]{AtmosphericPressureData}
Tiwari, A.
\newblock Performing analysis of meteorological data.
\newblock \url{https://github.com/Abhishek20182/Performing-Analysis-of-Meteorological/Data/blob/main/Project%201.ipynb}, 2021.

\bibitem[Zhang(2003)]{ZHANG2003159}
Zhang, G.
\newblock Time series forecasting using a hybrid arima and neural network model.
\newblock \emph{Neurocomputing}, 50:\penalty0 159--175, 2003.
\newblock ISSN 0925-2312.
\newblock \doi{https://doi.org/10.1016/S0925-2312(01)00702-0}.
\newblock URL \url{https://www.sciencedirect.com/science/article/pii/S0925231201007020}.

\bibitem[Zhao et~al.(2017)Zhao, Lu, Chen, Liu, and Wu]{7870510}
Zhao, B., Lu, H., Chen, S., Liu, J., and Wu, D.
\newblock Convolutional neural networks for time series classification.
\newblock \emph{Journal of Systems Engineering and Electronics}, 28\penalty0 (1):\penalty0 162--169, 2017.
\newblock \doi{10.21629/JSEE.2017.01.18}.

\bibitem[Zhiyenbayev(2023)]{CO2-PPM}
Zhiyenbayev, M.
\newblock Co2 parts per million (ppm) data.
\newblock \url{https://github.com/datasets/co2-ppm}, 2023.

\end{thebibliography}
\end{document}